\relax
\documentclass[letterpaper]{article} 
\usepackage{times}  
\usepackage{helvet} 
\usepackage{courier}  
\usepackage[hyphens]{url}  
\usepackage{graphicx} 
\usepackage{natbib}  
\usepackage{caption} 
\usepackage{algorithm}
\usepackage{subfigure}
\usepackage{amsmath}
\usepackage{amsfonts}
\usepackage{booktabs}
\title{Learning to Reweight Imaginary Transitions \\for Model-Based Reinforcement Learning}
\author{
	Wenzhen Huang,\textsuperscript{\rm 1,2}
	Qiyue Yin,\textsuperscript{\rm 1,2}
	Junge Zhang,\textsuperscript{\rm 1,2}
	Kaiqi Huang\textsuperscript{\rm 1,2,3}	
	\\
	\small \textsuperscript{\rm 1} School of Artificial Intelligence,  University of Chinese Academy of Sciences, Beijing, China \\
	\small \textsuperscript{\rm 2} CRISE, Institute of Automation,   Chinese Academy of Sciences, Beijing, China\\
	\small \textsuperscript{\rm 3} CAS Center for Excellence in Brain Science  and Intelligence Technology, Beijing, China\\
	
}

\begin{document}

\maketitle

\begin{abstract}
Model-based reinforcement learning (RL) is more sample efficient than model-free RL by using imaginary trajectories generated by the learned dynamics model. When the model is inaccurate or biased, imaginary trajectories may be deleterious for training the action-value and policy functions.
To alleviate such problem, this paper proposes to adaptively reweight the imaginary transitions, so as to reduce the negative effects of poorly generated trajectories. More specifically, we evaluate the effect of an imaginary transition by calculating the change of the loss computed on the real samples when we use the transition to train the action-value and policy functions. Based on this evaluation criterion, we construct the idea of reweighting each imaginary transition by a well-designed meta-gradient algorithm. Extensive experimental results demonstrate that our method outperforms state-of-the-art model-based and model-free RL algorithms on multiple tasks. Visualization of our changing weights further validates the necessity of utilizing reweight scheme.
\end{abstract}

\begin{figure*}
	\begin{center}
		\subfigure{
			\centering
			\includegraphics[width=0.92\linewidth]{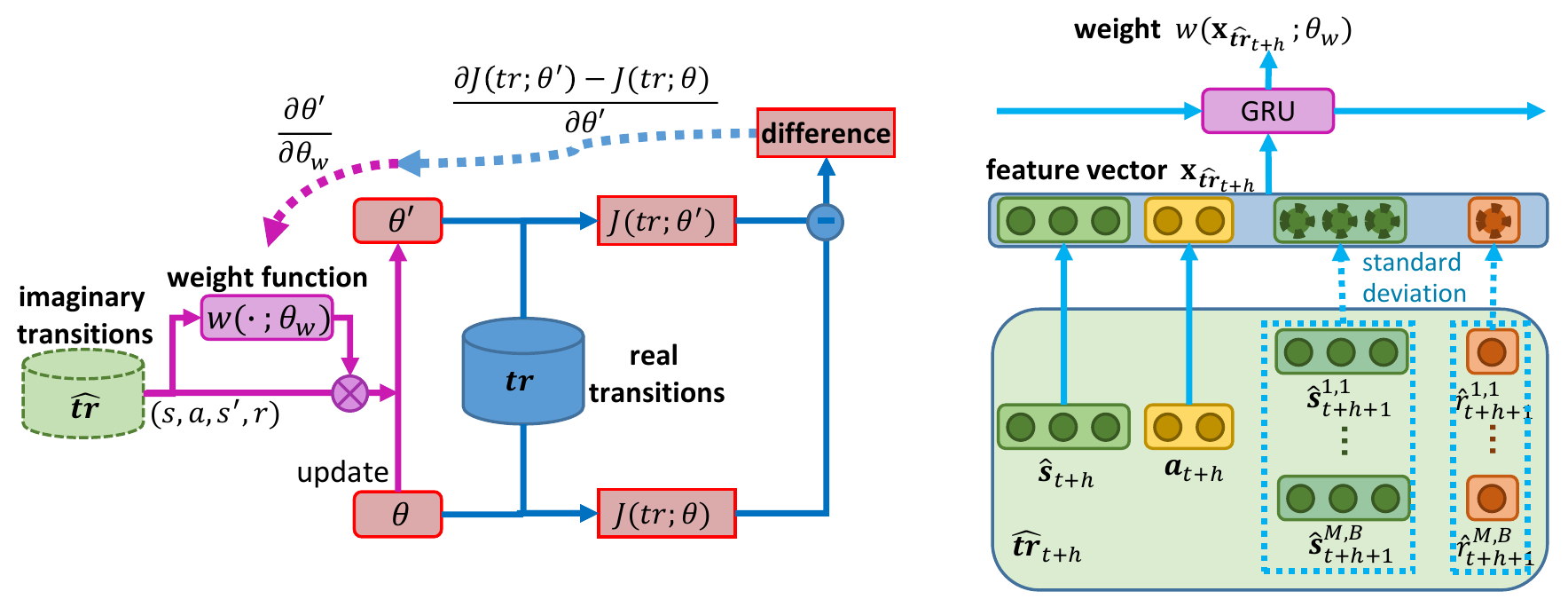}
		}%
	\end{center}
	\caption{\label{fig:total_framework}Training architecture (left) and Network architecture (right) for the weight function. We measure the negative effect of reweighted imaginary transitions through computing the difference of the losses computed on the real transitions before and after training with them, and minimize the difference to optimize the weight function by the chain rule.}
\end{figure*}
\section{Introduction}

Reinforcement learning (RL) algorithms are typically divided into two categories, i.e., model-free RL and model-based RL. The former directly learns the policy from the interactions with the environment, and has achieved impressive results in many areas, such as games~\citep{mnih2015human,silver2016mastering}. But these model-free algorithms are data-expensive to train, which limits their applications to simulated domains. Different from model-free approaches, model-based reinforcement learning algorithms learn an internal model of the real environment to generate imaginary data, perform online planning or do policy search, which holds promise to provide significantly lower sample complexity~\citep{luo2018algorithmic}.

Previously, model-based RL with linear or Bayesian models has obtained excellent performance on the simple low dimensional control problems~\citep{abbeel2006using,deisenroth2011pilco, levine2013guided, levine2014learning, levine2016end}. But these methods are hard to be applied to high-dimensional domains. Since neural network models can represent more complex transition functions, model-based RL with them can solve higher dimensional control problems~\citep{gal2016improving, depeweg2017learning, nagabandi2018neural}. However, learned high-capacity dynamics models ineluctably face predicting error, which results in the suboptimal performance and even catastrophic failures~\citep{deisenroth2011pilco}. 

Plenty of approaches have been proposed to alleviate the above problem. For example, \citep{chua2018deep} learns an ensemble of probabilistic models to mitigate the model error. \citep{clavera2018model} also learns the ensemble of models, and meta-trains a policy to adapt all the models so that the policy can be robust against model-bias. Among this line of research, a type of solution tries to tune model usage to reduce adverse effects of the imaginary data generated by inaccurate models, and promising results have been obtained. \citep{kalweit2017uncertainty} only uses imaginary trajectories in the case of high uncertainties of Q-function. \citep{heess2015learning} only uses imaginary data to compute policy gradients. \citep{janner2019trust} replaces model-generated rollouts begin from the initial state distribution with short model-generated rollouts branched from the real data. 

Above simple tuning schemes would result in that the generated data is always ignored in some training processes even it is completely accurate. Since samples with large prediction errors in the imaginary experience will lead to the value or policy function trained on it being inaccurate, adaptively filtering the samples with large prediction errors can reduce the performance degradation caused by the model bias. This makes a basic motivation of our study. However, the prediction error of an imaginary transition is difficult to obtain, because it is hard to decide a threshold of prediction error to determine whether the sample should be abandoned or not. For instance, when the value or policy function is very imprecise, even the samples with relatively large prediction errors can be used to optimize the function.

To handle above predication errors problem, we attempt to adaptively tune model usage through reweighting the imaginary samples according to their potential effect on training, which is totally different from previous model usage approaches. More specifically, we measure the effect through comparing the values of the optimization object (e.g, TD error) computed on the real samples before and after updating the functions using the imaginary transition. In this way, the filtering process can be taken as selecting an appropriate weight from {0, 1} for each imaginary sample based on its effect. To achieve this, we train a weight function to minimize adverse effects of the samples after they being reweighted using the function. The weight function outputs a weight in the range between 0 and 1 for each transition based on its features, like the uncertainty of the predicted next state in the transition. The effect of a reweighted sample can also be measured by the evaluation criterion mentioned previously. 

A main issue of using weight function lies in its optimization. Given a generated transition, a weight is predicted by the weight function and a weighted loss is accordingly calculated for updating parameters. Its effect is evaluated by the difference between the losses computed on the real transitions using the parameters before and after updating. As the loss is parameterized by the updated parameters and the update of parameters is parameterized by the output of the weight function, the function can be optimized through minimizing the difference using the chain rule. Our method can be considered as an instance of meta-gradient~\citep{xu2018meta,zheng2018learning,veeriah2019discovery}, a form of meta-learning~\citep{thrun1998learning,finn2017model,hospedales2020meta}, where the meta-learner is trained via gradients through the effect of the meta-parameters on a learner also trained via gradients~\citep{xu2018meta}.

To this end, we implement the algorithm by employing an ensemble of bootstrapped probabilistic neural networks and using Soft Actor-Critic~\citep{haarnoja2018soft, haarnoja2018soft1} to update the policy and action-value function. We name this implementation as Reweighted Probabilistic-Ensemble Soft-Actor-Critic (ReW-PE-SAC). Experimental results demonstrate that ReW-PE-SAC outperforms the state-of-the-art model-based and model-free deep RL algorithms on multiple benchmarking tasks. We also analyze the predicted weights on the samples generated with different schemes in different stages of the training process, which shows that the learned weight function can provide reasonable weights for different generated samples in different stages of the training process. In addition, the critic loss updated with the weighted samples is obviously smaller than the one updated with the unweighted samples. This means that the learned weight function can filter out the samples with adverse effects by decreasing their weights.

The main contributions of this work are:
\begin{itemize}
	\item We propose an effective tuning scheme of model usage through adaptively reweighting the imaginary transitions. 
	Different from the simple tuning schemes proposed by previous works, this theme can adaptively  filter generated samples with a certain degree of prediction error based on the precision of action-value and policy functions while maximizing the use of remaining generated samples.
	\item We use neural networks to predict the weight of each transition in the generated trajectories based on the well-designed features of the transitions and utilize meta-gradient method to optimize the weight network according to the above scheme. Thus, the learned weight network can be applied to new generated samples.
	\item Experimental results demonstrate that our method outperforms state-of-the-art model-based and model-free RL algorithms on multiple tasks.
\end{itemize}

\section{Approach}


\subsection{Notation}
Considering the standard reinforcement learning setting, an agent interacts with an environment in discrete time.
The environment is described by state space $\mathcal{S}$, action space $\mathcal{A}$,
reward function $r: \mathcal{S} \times \mathcal{A} \times \mathcal{S} \to \mathbb{R}$, state transition probabilities $p: \mathcal{S} \times \mathcal{S} \times \mathcal{A} \to [0, \infty)$, and a discount factor $\gamma \in (0, 1] $, where state transition probabilities $p(\mathbf{s}_t,\mathbf{a}_t,\mathbf{s}_{t+1})$ denotes the probability density of the next state $\mathbf{s}_{t+1}\in\mathcal{S}$ given the current state $\mathbf{s}_{t}\in\mathcal{S}$, action $\mathbf{a}_{t}\in\mathcal{A}$, and reward function $r(\mathbf{s}_t,\mathbf{a}_t,\mathbf{s}_{t+1})$ present the reward according to the transition.
At each time step $t$, the agent selects an action $\mathbf{a}_t$ according to the policy $\pi(\mathbf{a}_t|\mathbf{s}_t)$, and then receives the next state $\mathbf{s}_{t+1}$ and the reward $r_{t+1}$ from the environment.
The objective of standard reinforcement learning is to learn a policy of the agent to maximize the discount cumulative rewards.

\subsection{Overall Framework}
Model-based reinforcement learning approaches attempt to learn a dynamics model to simulate the real environment and utilize the model to make better decisions. 
In most cases, the learned model is imperfect and not all the transitions generated by it are accurate, which means the value and policy functions would be misled by the transitions with prediction errors.
Therefore, this paper proposes to adaptively reweight the generated transitions to minimize the negative effect of them for the training.

We train a weight function to minimize adverse effects of the transitions after they are reweighted.
Specifically, for a transition, the weight function outputs a weight.
The effect of a reweighted transition is measured by comparing the losses of value and policy functions computed on the real samples before and after the functions being updated by the reweighted transition.
As the loss before being updated is fixed, minimizing the adverse effect is equal to minimizing the loss after being updated.
This loss is parameterized by the updated parameters and the update of the parameters is parameterized by the weight function, thus we can optimize the function through minimizing the loss  after being updated by the chain rule.
The training process of weight function is shown in Figure~\ref{fig:total_framework}(left).

We employ an ensemble of bootstrapped probabilistic neural networks as the dynamics model, which can provide an estimated uncertainty for each generated transition.
The weight function can  predict the weights for the transitions more reasonably based on their estimated uncertainties.
We use Soft Actor-Critic~\citep{haarnoja2018soft, haarnoja2018soft1} to update the q-value and  policy functions, which is an off-policy RL algorithm so that we can use the old experience to evaluate the effect of the updated parameters.
We call this implementation as ReWeighted Probabilistic-Ensemble Soft-Actor-Critic (ReW-PE-SAC).

In the following, we would first present how to obtain the ensemble of networks, then describe the network architecture of the weight function, finally explain how to optimize the weight function.

\subsection{Dynamics Model}

In our method, the dynamics model is not only  required  to generate the transitions, but also needed to provide the other information that is useful for evaluating the weights of these transitions, like uncertainty.

In order to measure the uncertainties of generated transitions, we train an ensemble of $B$-many bootstrapped probabilistic models like~\citep{chua2018deep}. 
The $B$ models have the same architecture but different parameters $\mathbf{\theta}_b$ and training datasets $R_b$. 
Each dataset $R_b$ is generated by sampling with replacement $N$ times from the replay buffer $R$, where $N$ is equal to the size of $R$.
Each probabilistic model is a neural network that predicts the probability distribution of the next state $\mathbf{s}'$ based on the input state $\mathbf{s}$ and action $\mathbf{a}$.
The probability distribution is described by a Gaussian distribution, $\mathcal{N}(\mu_{\theta_b}(\mathbf{s}, \mathbf{a}), \Sigma_{\theta_b}(\mathbf{s}, \mathbf{a}))$.
The predicted next state is obtained by sampling from the Gaussian distribution, $\mathcal{N}(\mu_{\theta_b}(\mathbf{s}_n, \mathbf{a}_n), \Sigma_{\theta_b}(\mathbf{s}_n, \mathbf{a}_n))$.
Reward function $r(s,a,s'): \mathcal{S} \times \mathcal{A} \times \mathcal{S} \rightarrow \mathbb{R}$ is assumed as given in advance, like most works of literature related to model-based RL methods~\citep{wang2019benchmarking, clavera2018model, chua2018deep}.

Given a state $\mathbf{s}_t$ and an action sequence $\mathbf{a}_{t:t+H-1}=\{\mathbf{a}_t, \dots, \mathbf{a}_{t+H-1}\}$, the learned dynamics models can induce a distribution over the subsequent trajectories $\mathbf{s}_{t+1:t+H}$.
Based on $\mathbf{s}_t$ and $\mathbf{a}_t$, we use the ensemble of probabilistic models to induce $B$-many Gaussian distributions of the next state $\mathbf{s}_{t+1}$, and then sample $M$ states $\{\hat{\mathbf{s}}_{t+1}^{mb}\}_{m=1}^M$ from each  Gaussian distributions $\mathcal{N}(\mu_{\theta_b}(\mathbf{s}_t, \mathbf{a}_t), \Sigma_{\theta_b}(\mathbf{s}_t, \mathbf{a}_t))$.
The reward function is applied to the predicted next states to evaluate the reward of them, $\hat{r}_{t+1}^{mb} = r(\mathbf{s}_t,\mathbf{a}_t,\hat{\mathbf{s}}_{t+1}^{mb})$.
A state is randomly selected from the $M{\times}B$ predicted states $\{\hat{\mathbf{s}}_{t+1}^{mb}\}_ {m=1,b=1}^{M,B}$ as the next input $\hat{\mathbf{s}}_{t+1}$.
Then the selected state $\hat{\mathbf{s}}_{t+1}$ and the action $\mathbf{a}_{t+1}$ are used to generate the subsequent $M{\times}B$ states.
In this way, we can get a transition set $\hat{\mathbf{tr}}_{k} = \{(\hat{\mathbf{s}}_{t+k}, \mathbf{a}_{t+k}, \hat{r}_{t+k}^{b,m}, \hat{\mathbf{s}}_{t+k+1}^{b,m})\}_{m=1,b=1}^{M,B}$ for each time-step $t+k, k=0,...,H-1$.

\subsection{Weight Prediction Network}

Estimating the weight on a single generated transition $({\mathbf{s}}, \mathbf{a}, \hat{r}, \hat{\mathbf{s}'})$ is difficult, because we cannot obtain any information about the prediction accuracy of $\hat{r}$ and $\hat{\mathbf{s}'}$ from the single transition.
Thus, we estimate the weight on the transition set\\ $\hat{\mathbf{tr}}_{k} = \{(\hat{\mathbf{s}}, \mathbf{a}, \hat{r}^{b,m}, \hat{\mathbf{s}'}^{b,m})\}_{m=1,b=1}^{M,B}$ generated by the ensemble of probabilistic models for the input $(\mathbf{s}, \mathbf{a})$ instead of the single transition.

The weight function $\mathrm{w}(\mathbf{x}_\mathbf{tr};\theta_w): \mathbb{R}^D\rightarrow(0, 1)$ is approximated by a neural network with parameters $\theta_w$, where $\mathbf{x}_\mathbf{tr}$ represents the feature vector of a generated transition set $\mathbf{tr} = \{(\mathbf{s}, \mathbf{a}, \hat{r}^{b,m}, \hat{\mathbf{s}'}^{b,m})\}_{m=1,b=1}^{M,B}$.
The feature vector $\mathbf{x}_\mathbf{tr}$ is composed of the states $\mathbf{s}$, the actions $\mathbf{a}$, the uncertainty on the predicted reward $\hat{r}$ and the uncertainties on each dimension of the predicted next state $\hat{\mathbf{s}'}$.
The uncertainties are approximated by computing the standard deviation of rewards and the next states $\{(\hat{r}^{b,m}, \hat{\mathbf{s}'}^{b,m})\}_{m=1,b=1}^{M,B}$.
The uncertainties imply the credibility of the generated transition $\hat{\mathbf{tr}}$, while the inputted state and action uniquely identify the transitions.
In practice, we find the latter one enables the weight function to make a better prediction.
To avoid the large disparities of different features, the feature vectors are normalized for each dimension before they are fed to the weight network.

It is obvious that the credibility of $\hat{\mathbf{tr}}_{k}$ is related to the ones of its predecessors $\{\hat{\mathbf{tr}}_j\}_{j<k}$, due to that modeling errors in dynamics are accumulated with time-steps.
Thus we select Gated Recurrent Units (GRU)~\citep{cho2014learning} to integrate the features of the predecessors.
The network architecture of weight function is shown in Figure~\ref{fig:total_framework}(right).

\begin{algorithm}[htb] 
	
	\caption{Reweighted Probabilistic-Ensemble Soft-Actor-Critic (ReW-PE-SAC)\label{alg:meta_reweight}} 
	\textbf{Input:} the learning rate $\mu$ of $\theta_q$, $\theta_\pi$, $\alpha$ and the learning rate $\mu_w$ of $\theta_w$
	
	\textbf{Init:} initialize parameters $\theta_q$, $\theta_\pi$, $\alpha$, $\theta_w$ and replay buffer $R\leftarrow\emptyset$
	
	\textbf{for} $t=1,2,\dots,N$ \textbf{do}
	
	\quad Interact with the real environment based on the current policy, and add the transitions to replay buffer $R$
	
	\quad Train the dynamics models using replay buffer $R$
	
	\quad 
	
	\quad // Training the weight function
	
	\quad Generate imaginary transitions $\{\hat{\mathbf{tr}}_h^i\}_{i=1,h=1}^{N_e, H}$
	
	\quad Sample real transitions $\mathbf{tr}$ from replay buffer $R$
	

	\quad Update $\theta_q, \theta_\pi$ to $\theta_q', \theta_\pi'$ by Equation~\ref{equ:update_q_with_imaginary} using $\{\hat{\mathbf{tr}}_h^i\}_{i=1,h=1}^{N_e, H}$

	\quad Compute the meta objective $J_{meta}$ by Equation~\ref{equ:meta_objective} on $\mathbf{tr}$
	
	\quad Approximate the gradient of $\nabla_{\theta_w}J_{meta}$ by Equation~\ref{equ:chain_rule}
	
	\quad Update $\theta_w' \leftarrow \theta_w - \mu_w \nabla_{\theta_w}J_{meta}$
	
	\quad 
	
	\quad // Update value and policy network 
	
	\quad Generate imaginary dataset $\{\hat{\mathbf{tr}}_h^i\}_{i=1,h=1}^{N_t, H}$
	
	\quad \textbf{for} $k=1,2,\dots,K$ \textbf{do}
	
	\quad \quad Update $\theta_q$, $\theta_\pi$ by Equation~\ref{equ:train_q_with_imaginary} on the reweighted imaginary samples
	
	\quad \textbf{end for}
	
	\quad Sample real transitions $\mathbf{tr}$ to update $\theta_q$, $\theta_\pi$
	
	\textbf{end for}
\end{algorithm}

\subsection{Training the Weight Function}
\label{sec:learning_weight_function}

This section will show how to train the weight function so that it can predict appropriate weights for imaginary transitions to minimize their adverse effect.

The training of weight function can be split into two steps,
evaluating the potential effects of the reweighted transitions and optimizing the weight function through minimizing the negative effects by the chain rule.
We focus on the effects of the action-value and policy functions, and update the weight function through minimizing the effects of a mini-batch of imaginary transitions in each iteration.

For the first step, we sample $N_e$ real states $\{\mathbf{s}_{t_i}^i\}_{i=1}^{N_e}$ and the corresponding real action sequences $\{\mathbf{a}_{{t_i}:{t_i}+H-1}^i\}_{i=1}^{N_e}$ from the replay buffer $\mathbb{D}$ to generate the imaginary transitions $\{\hat{\mathbf{tr}}_h^i\}_{i=1,h=1}^{N_e, H}$, where $H$ is the planning horizon.
Then we compute the weights of imaginary transitions $\mathrm{w}(\mathbf{x}_{\hat{\mathbf{tr}}_h^i};\theta_w)$ and update the parameters of Q-network and policy network, $\theta_q$ and $\theta_\pi$, with the reweighted losses of these imaginary samples:
\begin{equation}\label{equ:update_q_with_imaginary}
\begin{split}
\theta_q' = \theta_q - \mu \frac{\partial \sum_{i,h} \mathrm{w}(\mathbf{x}_{\hat{\mathbf{tr}}_h^i};\theta_w)  J_{Q}(\hat{\mathbf{tr}}_h^i;\theta_q)}{\partial \theta_q}, \\
\theta_\pi' = \theta_\pi - \mu \frac{\partial \sum_{i,h} \mathrm{w}(\mathbf{x}_{\hat{\mathbf{tr}}_h^i};\theta_w)  J_{\pi}(\hat{\mathbf{tr}}_h^i;\theta_\pi)}{\partial \theta_\pi},
\end{split}
\end{equation}
where $\mu$ is the learning rate of $\theta_q$ and $\theta_\pi$. $J_{Q}$ and $J_{\pi}$ are the soft Bellman residual and the KL-divergence between the policy and the exponential of the soft Q-function~\citep{haarnoja2018soft, haarnoja2018soft1}, respectively.
For a transition set $\mathbf{tr}$, $J_{Q}$ and $J_{\pi}$ are computed by
\begin{align}\label{equ:soft_bellman_residual}
J_Q(\mathbf{tr};\theta_q)=&\sum_{(\mathbf{s}, \mathbf{a}, \mathrm{r}, \mathbf{s}')\in\mathbf{tr}} 
\frac{1}{2}\big\{ 
Q(\mathbf{s}, \mathbf{a}; \theta_q)-
 \notag \\
&[\mathrm{r} + \gamma(Q(\mathbf{s}', {\mathbf{a}'};\bar{\theta}_q) -\alpha {\rm log}\pi({\mathbf{a}}'|\mathbf{s}'))] \big\}^2,
\end{align}
\begin{equation}\label{equ:sac_policy_update}
\mathit{J}_{\pi}(\mathbf{tr};\theta_\pi)=\sum_{(\mathbf{s}, \mathbf{a}, \mathrm{r}, \mathbf{s}')\in\mathbf{tr}}
\alpha{\rm log} (\pi(\hat{\mathbf{a}}|\mathbf{s}_t; \theta_\pi))
-Q(\mathbf{s}_t, \hat{\mathbf{a}}; \theta_q),
\end{equation}
where $\bar{\theta}_q$ is the parameters of target Q-network, and $\alpha$ is the temperature parameter.

For the second step, we sample $N_v$ real transitions from the replay buffer $\mathbb{D}$, combined them into a set $\mathbf{tr}$, and compute the losses of q-value and policy functions on them with the updated parameters $\theta_q'$ and $\theta_\pi'$,
\begin{equation}\label{equ:meta_objective}
J_{Q}(\mathbf{tr};\theta_q') + J_{\pi}(\mathbf{tr};\theta_\pi').
\end{equation}
The gradient of the parameters of weight function $\theta_w$ is computed through the chain rule,
\begin{align}
& \frac{\partial J_{Q}(\mathbf{tr};\theta_q') + J_{\pi}(\mathbf{tr};\theta_\pi')}{\partial \theta_w} \notag\\
=& \frac{\partial  J_{Q}(\mathbf{tr};\theta_q')}{\partial \theta_q'}\frac{\partial \theta_q'}{\partial \theta_w} + \frac{ J_{\pi}(\mathbf{tr};\theta_{\pi}')}{\partial \theta_\pi'}\frac{\partial \theta_\pi'}{\partial \theta_w} \notag \\
=& -\mu \sum_{h,i}\big[ 
(\frac{\partial J_{Q}(\hat{\mathbf{tr}}_{h}^i;\theta_q)}{\partial \theta_q})^T \frac{\partial J_{Q}(\mathbf{tr};\theta_q')}{\partial \theta_q'} \notag \\
&
+ (\frac{\partial J_{\pi}(\hat{\mathbf{tr}}_{h}^i;\theta_\pi)}{\partial \theta_\pi})^T \frac{\partial  J_{\pi}(\mathbf{tr};\theta_\pi')}{\partial \theta_\pi'}
\big]
\frac{\partial \mathrm{w}(\mathbf{x}_{\hat{\mathbf{tr}}_h^i};\theta_w)}{\partial \theta_w} \label{equ:chain_rule}
\end{align}
Once the gradient is obtained, the parameters $\theta_w$ can be updated by any optimization algorithm.

We alternately optimize the q-value, policy functions, and the weight function, so that the latter one can adaptively adjust the weights of imaginary transitions along with the change of the precision of the former ones.
We sample $N_t$ real states and the corresponding action sequences with an explore policy $\pi_e$ which is obtained by changing the temperature parameter of current policy from $\alpha$ to $\lambda_e \alpha$ ($\lambda_e$ is set to 10 in this paper).
A larger temperature parameter is conducive to generating diverse transitions.
Based on the sampled state and  action sequences, we utilize the dynamics model to generate imaginary transitions $\{\hat{\mathbf{tr}}_h^i\}_{i=1,h=1}^{N_t, H}$ and use the weight function to reweight them.
The gradients of q-value and policy functions are computed by 
\begin{equation}\label{equ:train_q_with_imaginary}
\begin{split}
\nabla\theta_q =  \frac{\partial \sum_{i,h} \mathrm{w}(\mathbf{x}_{\hat{\mathbf{tr}}_h^i};\theta_w)  J_{Q}(\hat{\mathbf{tr}}_{h}^i;\theta_q)}{\partial \theta_q}, \\
\nabla\theta_\pi =  \frac{\partial \sum_{i,h} \mathrm{w}(\mathbf{x}_{\hat{\mathbf{tr}}_h^i};\theta_w)  J_{\pi}(\hat{\mathbf{tr}}_{h}^i;\theta_\pi)}{\partial \theta_\pi}.
\end{split}
\end{equation}
We use Adam to update the parameters $\theta_q$ and $\theta_\pi$.
The temperature parameter $\alpha$ is optimized based on the generated transition sets without being reweighted.

\begin{table*}[t]
	\tiny
	\begin{center}
		
		\begin{tabular}{lcccccc}
			\toprule
			& Ant & HalfCheetah & Hopper & Slimhumanoid & Swimmer & Walker2d \\
			\midrule
			
			ME-TRPO    & 282.2$\pm$18.0 & 2283.7$\pm$900.4 & 1272.5$\pm$500.9 & -154.9$\pm$534.3 & 30.1$\pm$9.7 & -1609.3$\pm$657.5\\
			MB-MPO & 705.8 $\pm$ 147.2 & 3639.0$\pm$1185.8 & 333.2$\pm$1189.7 & 674.4$\pm$982.2 & \textbf{85.0$\pm$98.9} & -1545.9$\pm$216.5\\
			PETS & 1165.5$\pm$226.9 & 2795.3$\pm$879.9 & 1125.0$\pm$679.6 & 1472.4$\pm$738.3 & 22.1$\pm$25.2 & 260.2$\pm$536.9\\
			POPLIN & 2330.1$\pm$320.9 & 4235.0$\pm$1133.0 & 2055.2$\pm$613.8 & -245.7$\pm$141.9 & 37.1$\pm$4.6 & 597.0$\pm$478.8\\
			MBPO & 4332.5$\pm$1277.6 & 10758.9$\pm$1413.7 & \textbf{3279.8$\pm$455.0} & 2950.4$\pm$819.1 & 26.3$\pm$13.3 & 4154.7$\pm$846.1\\
			
			\midrule
			TD3     & 956.1$\pm$66.9 & 3614.3$\pm$82.1 & 2245.3$\pm$232.4 & 1319.1$\pm$1246.1 & 40.4$\pm$8.3 & -73.8$\pm$769.0\\
			SAC-200k    &922.0$\pm$283.0 & 6129.3$\pm$775.7 & 2365.1$\pm$193.4 & 1891.6$\pm$379.2 & 49.7$\pm$5.8 & 1642.7$\pm$606.9 \\
			\midrule
			

			w.o reweighting & 4033.5$\pm$1480.5 & \textbf{11854.3$\pm$102.8} & 2202.6$\pm$363.5 & 1436.8$\pm$490.8 & 26.6$\pm$25.4 & 2673.8$\pm$2264.8 \\
			Our Method & \textbf{4614.4$\pm$931.1} & 9779.8$\pm$546.6 & 2824.0$\pm$159.9 & \textbf{11755.9$\pm$11152.2} & \textbf{82.2$\pm$33.4} & \textbf{4961.9$\pm$457.8}\\
			
			
			\midrule
			\midrule
			SAC-1000k  & 4994.9$\pm$719.5 & 10283.8$\pm$648.4 & 2990.3$\pm$214.3 & 29122.5$\pm$11129.0 & 86.8$\pm$6.4 & 5094.0$\pm$1371.3 \\
			
			\bottomrule
		\end{tabular}
		
	\end{center}
	\caption{
		Final performance on the six environments. All the algorithms are run for 200k time-steps (except SAC-1000k).  The results are shown with the mean and standard deviation averaged and a window size of 5000 times-steps.
	}
	\label{tab:performance}
\end{table*}

The complete algorithm is shown in Alg.~\ref{alg:meta_reweight}.
In our algorithm, the real transitions are not only used to train the dynamics models, but also used to train the action-value and policy networks.
The real samples can avoid too large prediction errors of the action-value function.
When the predicted weights of generated samples are too low, the real samples can prevent algorithm from being in stagnation behavior.

\begin{figure*}
	\centering
		\subfigure[Ant]{
			\centering
			\includegraphics[width=0.3\linewidth]{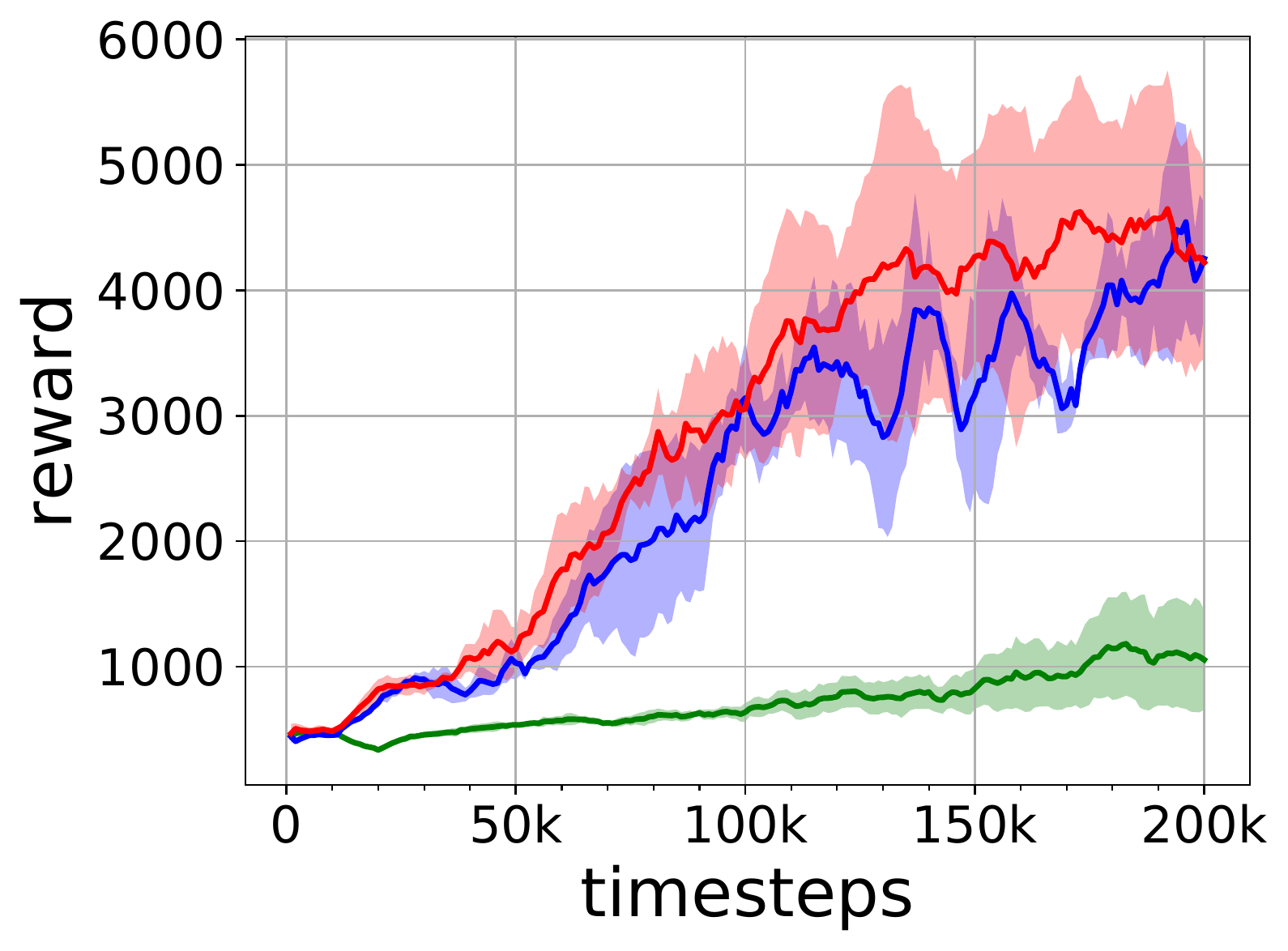}
		}%
		\subfigure[HalfCheetah]{
			\centering
			\includegraphics[width=0.3\linewidth]{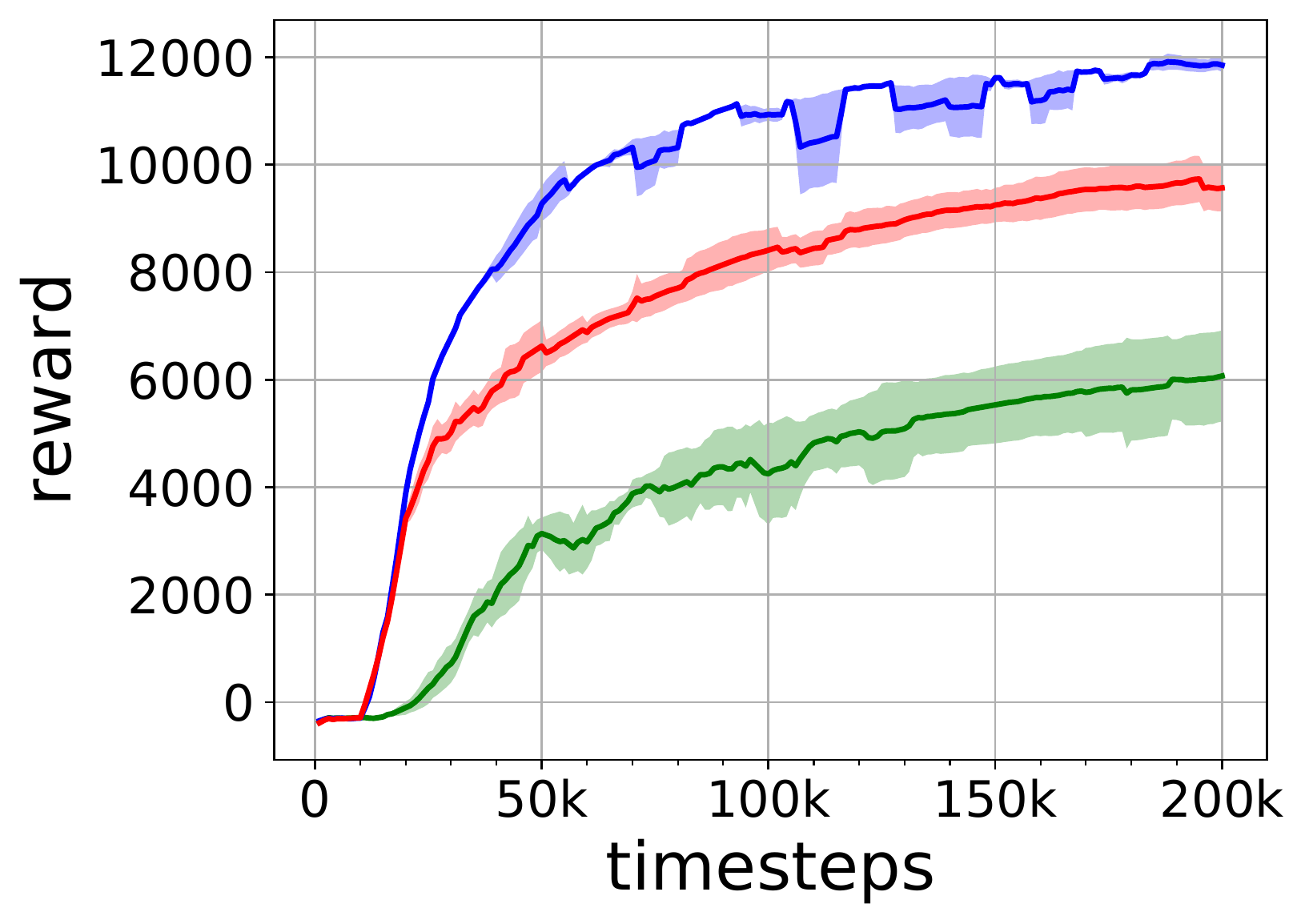}
		}%
		\subfigure[Hopper]{
			\centering
			\includegraphics[width=0.3\linewidth]{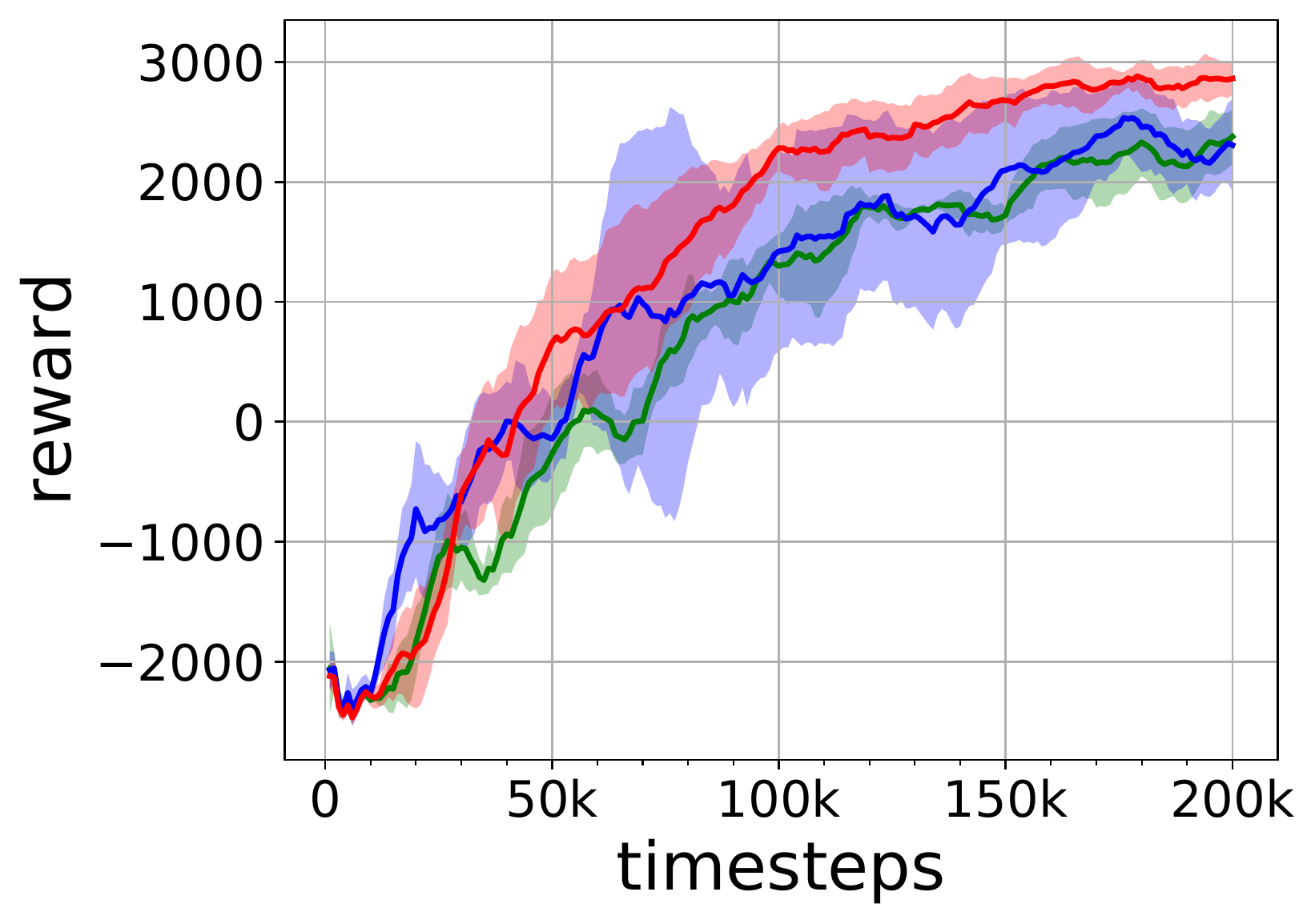}
		}%
	\\
		\subfigure[Slimhumanoid]{
			\centering
			\includegraphics[width=0.3\linewidth]{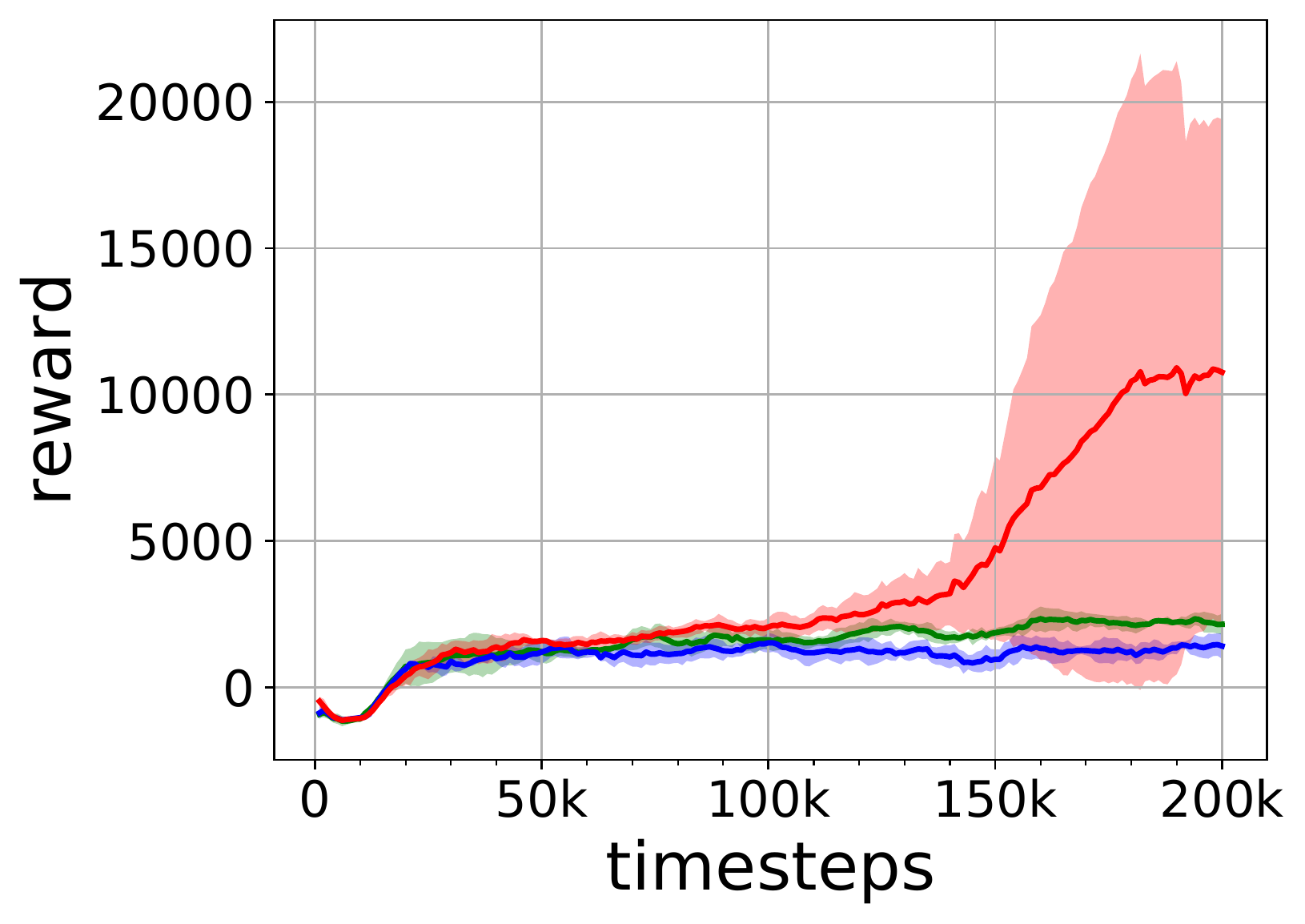}
		}%
		\subfigure[Swimmer\label{fig:performance_swimmer}]{
			\centering
			\includegraphics[width=0.3\linewidth]{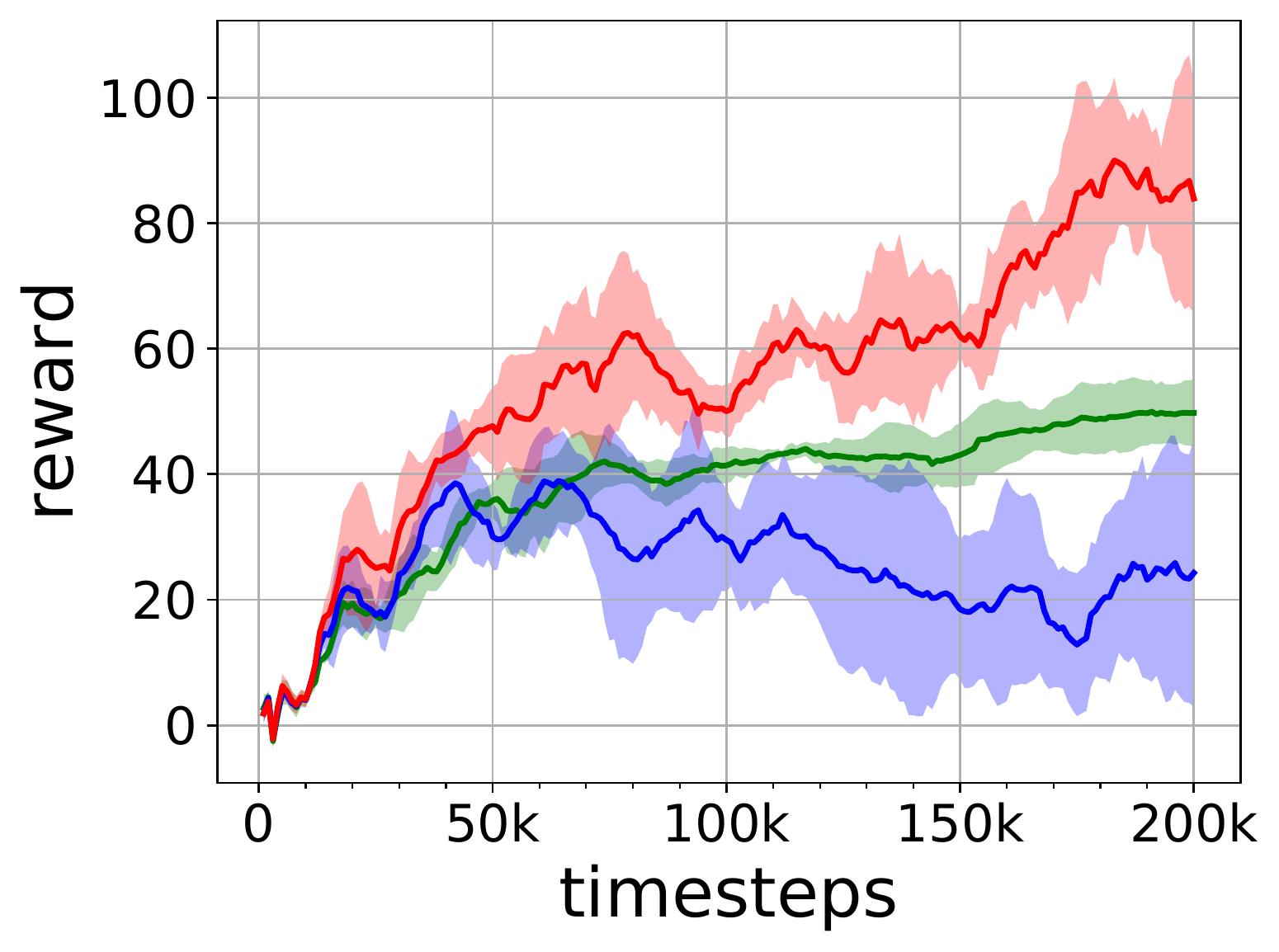}
		}%
		\subfigure[Walker2D]{
			\centering
			\includegraphics[width=0.3\linewidth]{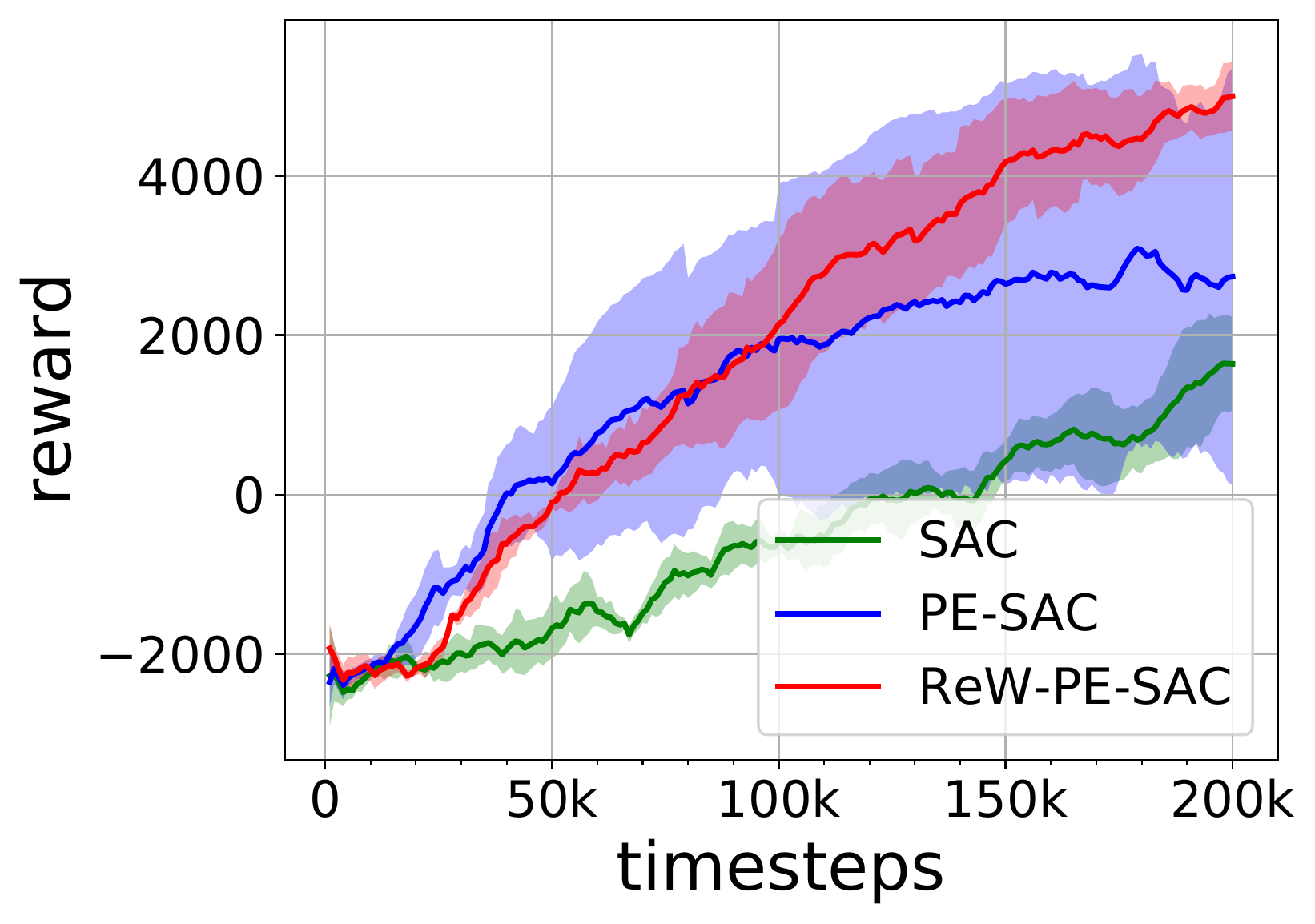}
		}%
	\caption{\label{fig:performance} Learning curves for different tasks and algorithms. All the algorithms are run for 200k time-steps with 8 random seeds.}
\end{figure*}

\section{Experiments}

In this section, we evaluate our algorithm on six complex continuous control tasks from the model-based RL benchmark~\citep{wang2019benchmarking}, which is modified from the OpenAI gym benchmark suite~\citep{brockman2016openai}.
The six tasks are Ant, HalfCheetah, Hopper, SlimHumanoid, Swimmer-v0, and Walker2D, whose horizon length is fixed to 1000. 
The network architecture and training hyperparameters are given in the appendix.
First, we compare ReW-PE-SAC on the benchmark against state-of-the-art model-free and model-based approaches.
Then, we show the differences of the q-value losses with and without reweighting method.
Next, we evaluate the robustness of our algorithm to imperfect dynamics model.
Finally, we analyze the relation between the learned weights and the factors of the training iterations, the planning horizon, and the explore policy.

\begin{figure}
	\centering
	\subfigure[Ant]{
		\centering
		\includegraphics[width=0.33\linewidth]{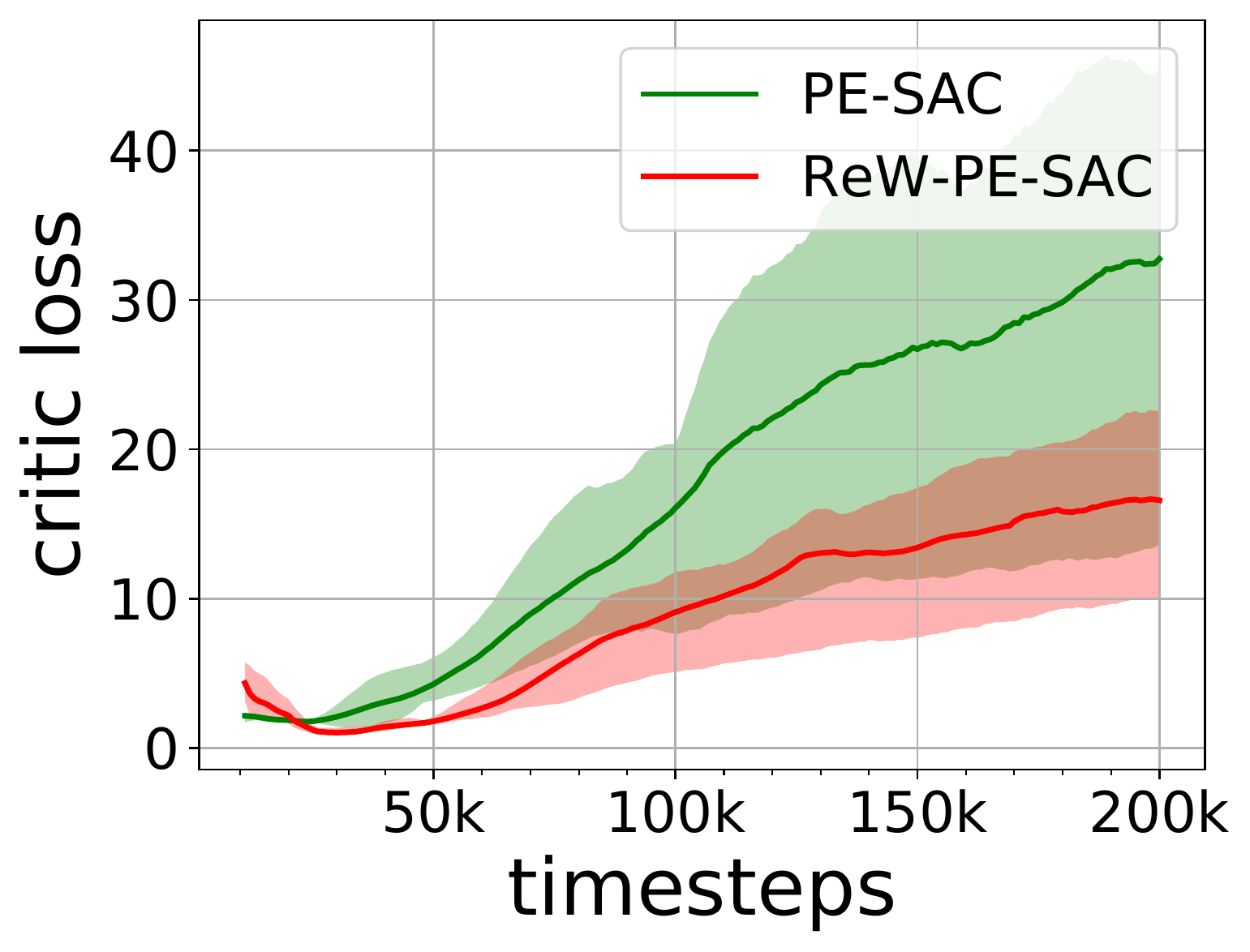}
	}%
	\subfigure[HalfCheetah]{
		\centering
		\includegraphics[width=0.33\linewidth]{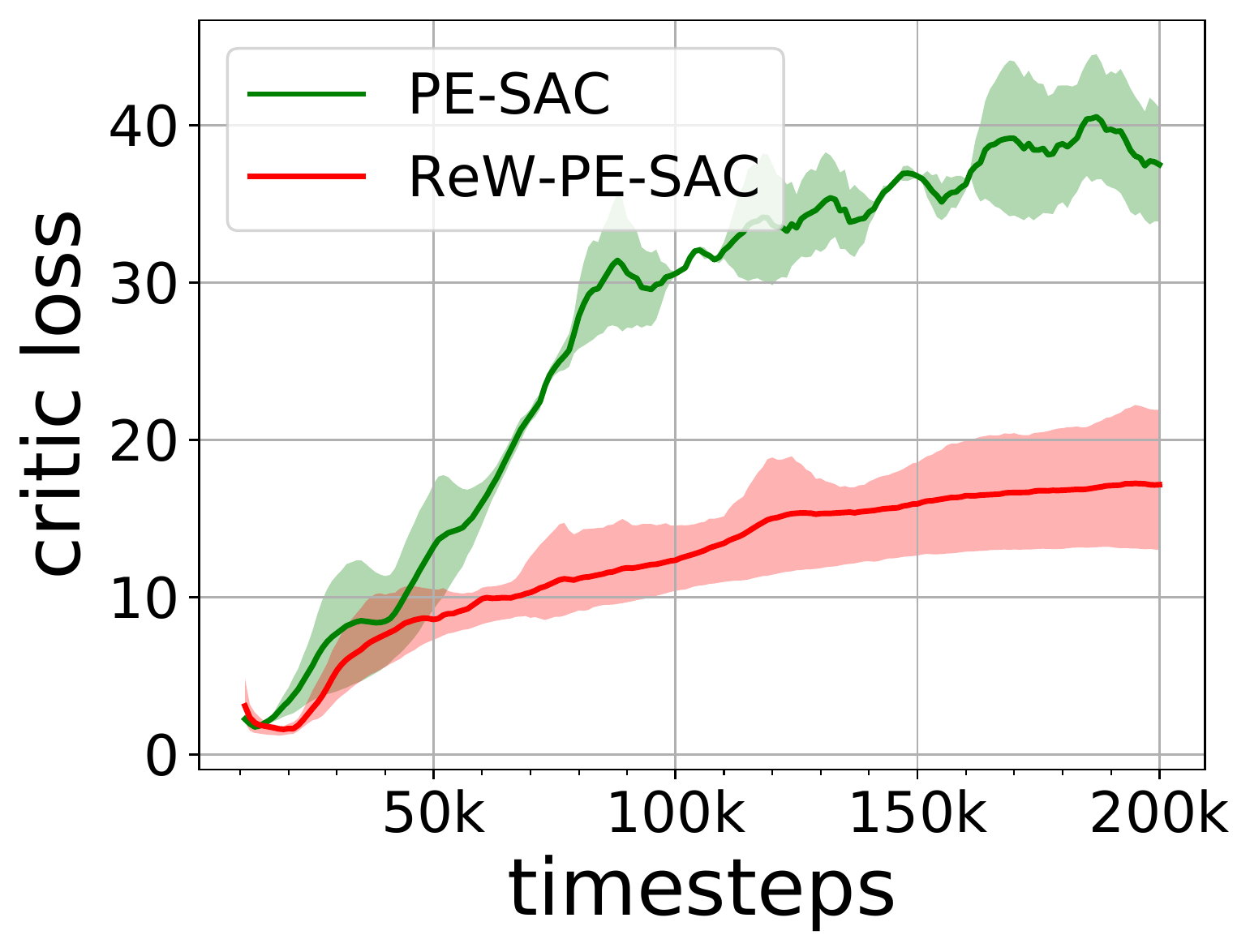}
	}%
	\\
	\subfigure[SlimHumanoid]{
		\centering
		\includegraphics[width=0.33\linewidth]{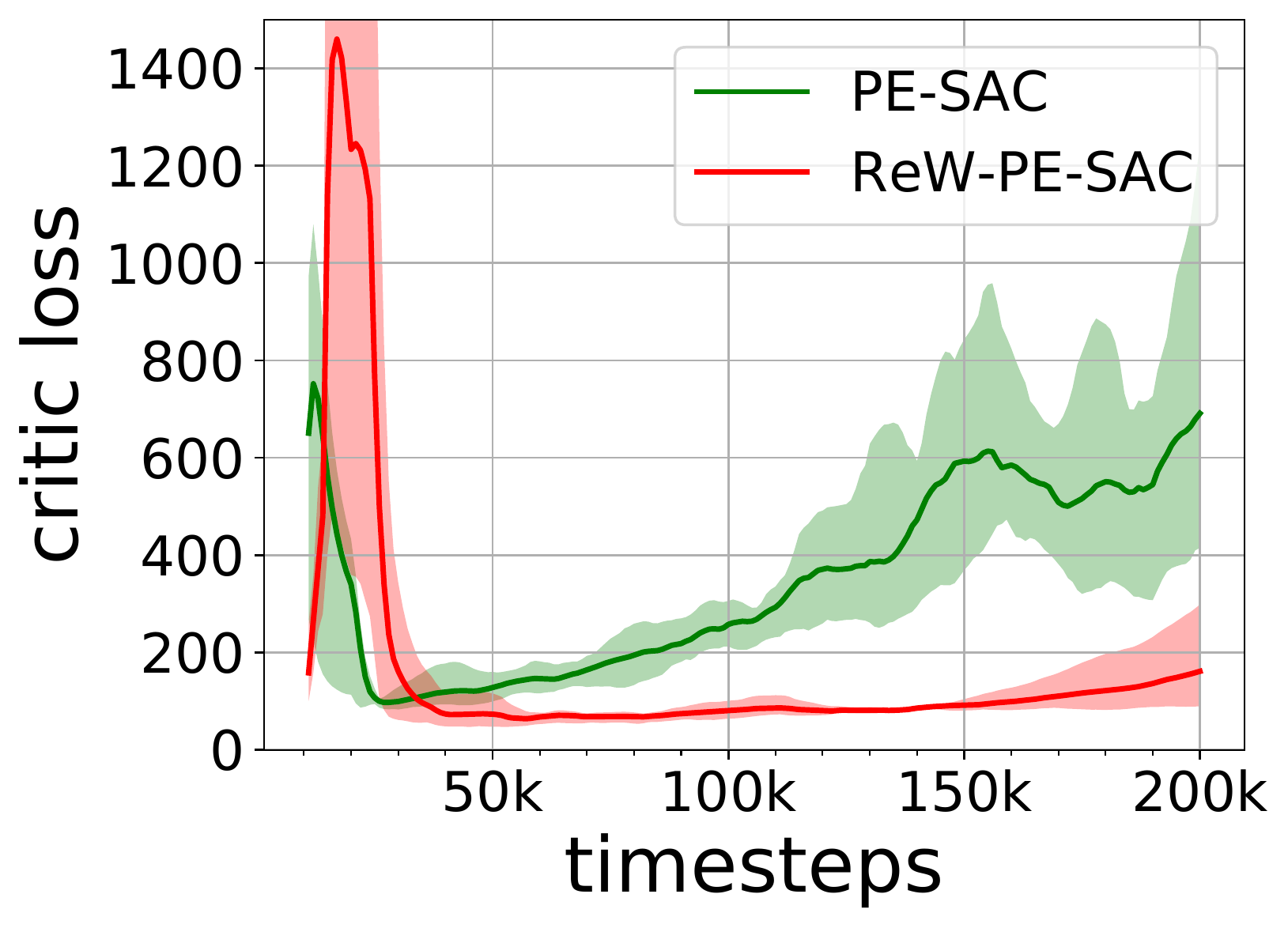}
	}%
	\subfigure[Swimmer]{
		\centering
		\includegraphics[width=0.33\linewidth]{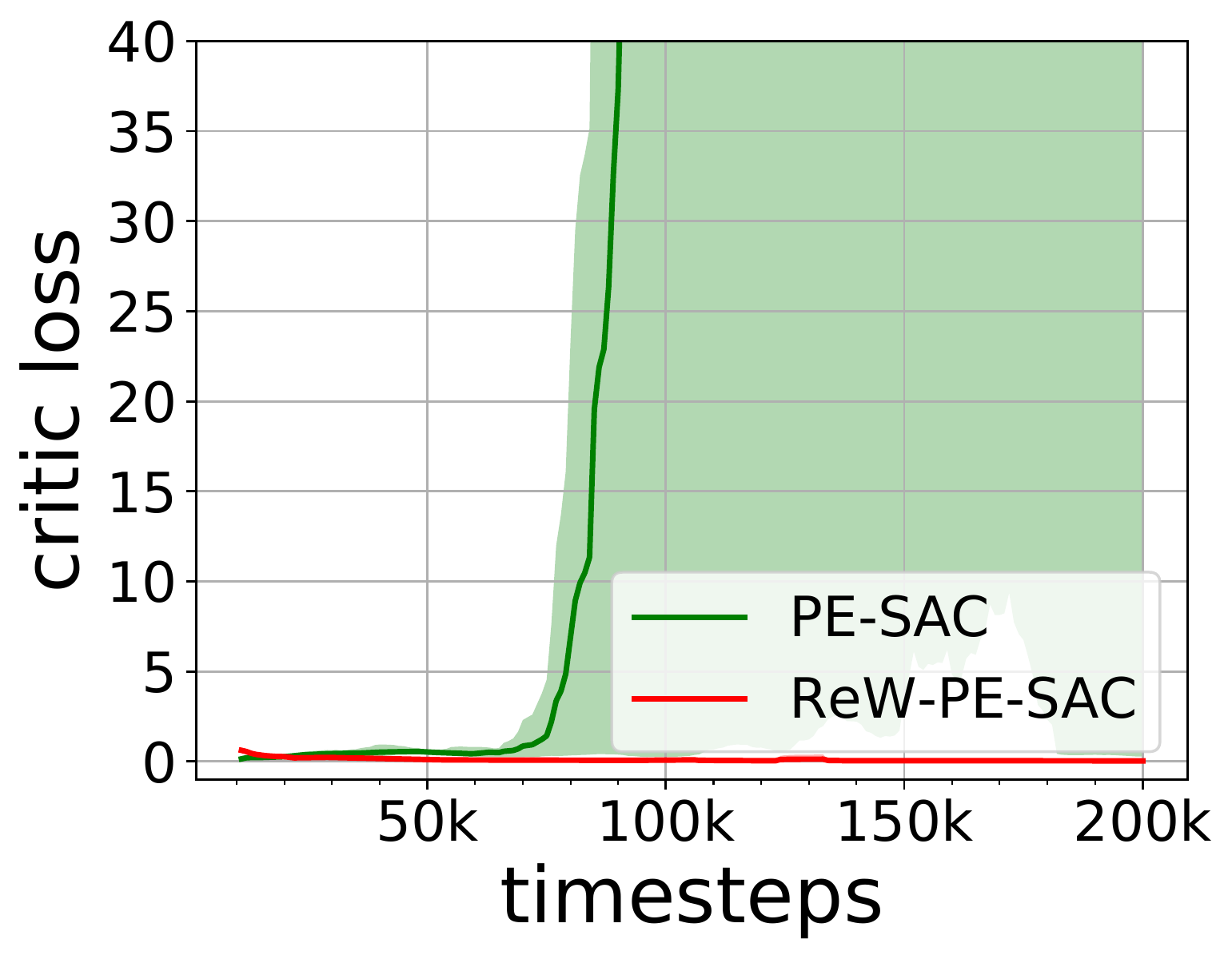}
	}%
	\caption{Critic losses in the cases with and without reweighting. The x-axis corresponds to time-step. The y-axis corresponds to average critic loss over 1 episode (1000 time-steps).}
	\label{fig:value_loss}
\end{figure}

\subsection{Comparison with State of the Art}
\label{sec:comparison}

We compare ReW-PE-SAC with state-of-the-art model-free and model-based RL methods, including SAC~\citep{haarnoja2018soft, haarnoja2018soft1}\footnote{We select the PyTorch implement of soft actor-critic in https://github.com/pranz24/pytorch-soft-actor-critic to evaluate the performance. This implement includes using double-Q network, ignoring the artificial terminal signal and other tricks, so the performance is better than the one reported in~\citep{wang2019benchmarking}.}, TD3~\citep{fujimoto2018addressing}, ME-TRPO~\citep{kurutach2018model}, MB-MPO\citep{clavera2018model}, PETS~\citep{chua2018deep},  MBPO~\citep{janner2019trust} and POPLIN~\citep{wang2019exploring}.
We reproduce results from \citep{wang2019benchmarking, janner2019trust} and additionally run MBPO on the tasks of Slimhumanoid and Swimmer as the according experimental results are absent.
We run our method ReW-PE-SAC for $200,000$ time-steps with $8$ random seeds.
To evaluate our reweighting mechanism, we also run PE-SAC on these six tasks which does not learn the weight function and directly use the imaginary transitions to train the policy and value networks.
To measure the sample efficiency of ReW-PE-SAC, we additionally run SAC $1,000,000$ time-steps on each task.
The results are summarized in Table~\ref{tab:performance}, and the learning curves of SAC and our methods with or without reweighting are plotted in Figure~\ref{fig:performance}.

As shown in Table~\ref{tab:performance}, ReW-PE-SAC achieves better performance compared with all other state-of-the-art algorithms except MBPO running with $200,000$ time-steps in all the environments.
Especially in the environments of Ant, Hopper, Swimmer and Walker2d, the performance of ReW-PE-SAC is comparable to the one of SAC running with $1,000,000$ time-steps, which demonstrates that ReW-PE-SAC has good sample efficiency.
Compared with MBPO, ReW-PE-SAC is better on four environments and is slightly weaker in the tasks of HalfCheetah and Hopper.

Comparing the results of our methods with and without reweighting, ReW-PE-SAC and PE-SAC,
the performance with reweighting is obviously higher on the most of the environments. This demonstrates that the learned weight function can provide appropriate weights to facilitate training a better policy.
The performance gap of ReW-PE-SAC and PE-SAC on the environment of HalfCheetah is probably caused by that the weight function is overcautious, and the weights provided by it are too low.

From Figure~\ref{fig:performance}(a,d), we find our method has a large performance variance in the tasks of Ant and Slimhumanoid. The most likely reason is that our method utilizes the collected transitions to evaluate the effect of imaginary transitions, while the number of collected transitions is insufficient for some tasks. This induces that the weights of some valid imaginary transitions could be underestimated, and then the learned policy would be relatively poor due to the lack of these valid transitions. We will consider constructing a more reasonable validation set in future work.

\subsection{The Critic Losses of PE-SAC and ReW-PE-SAC}
\label{sec:qvalue_losses}

In this section, we compare the critic losses in cases with and without reweighting. We run the algorithms of PE-SAC and ReW-PE-SAC on the tasks of Ant, HalfCheetah, SlimHumanoid and Swimmer, and record the average critic losses of real samples in every episode. The minimum, maximum and mean of the losses in the same time-step are plotted in Figure~\ref{fig:value_loss}.

As shown in the figure, ReW-PE-SAC can maintain lower critic losses than PE-SAC and prevent abnormal large losses.
Combined with the learning curve for the task of Swimmer (shown in Figure~\ref{fig:performance_swimmer}), we find the performance of PE-SAC is falling after about $70,000$ time-steps while the critic loss is also increasing sharply at around this time.
So maintaining lower losses has contributed to improve the performance in most cases.
The only exception is the task of HalfCheetah, in which the lower critic losses have not resulted in higher performance.
The most likely reason is that an imprecise Q-value function is enough to train a good policy.

\subsection{Robustness to Imperfect Dynamics Model}
\label{sec:diff_dynamic_model}

\begin{figure}[h]
	\centering
	\includegraphics[width=0.5\linewidth]{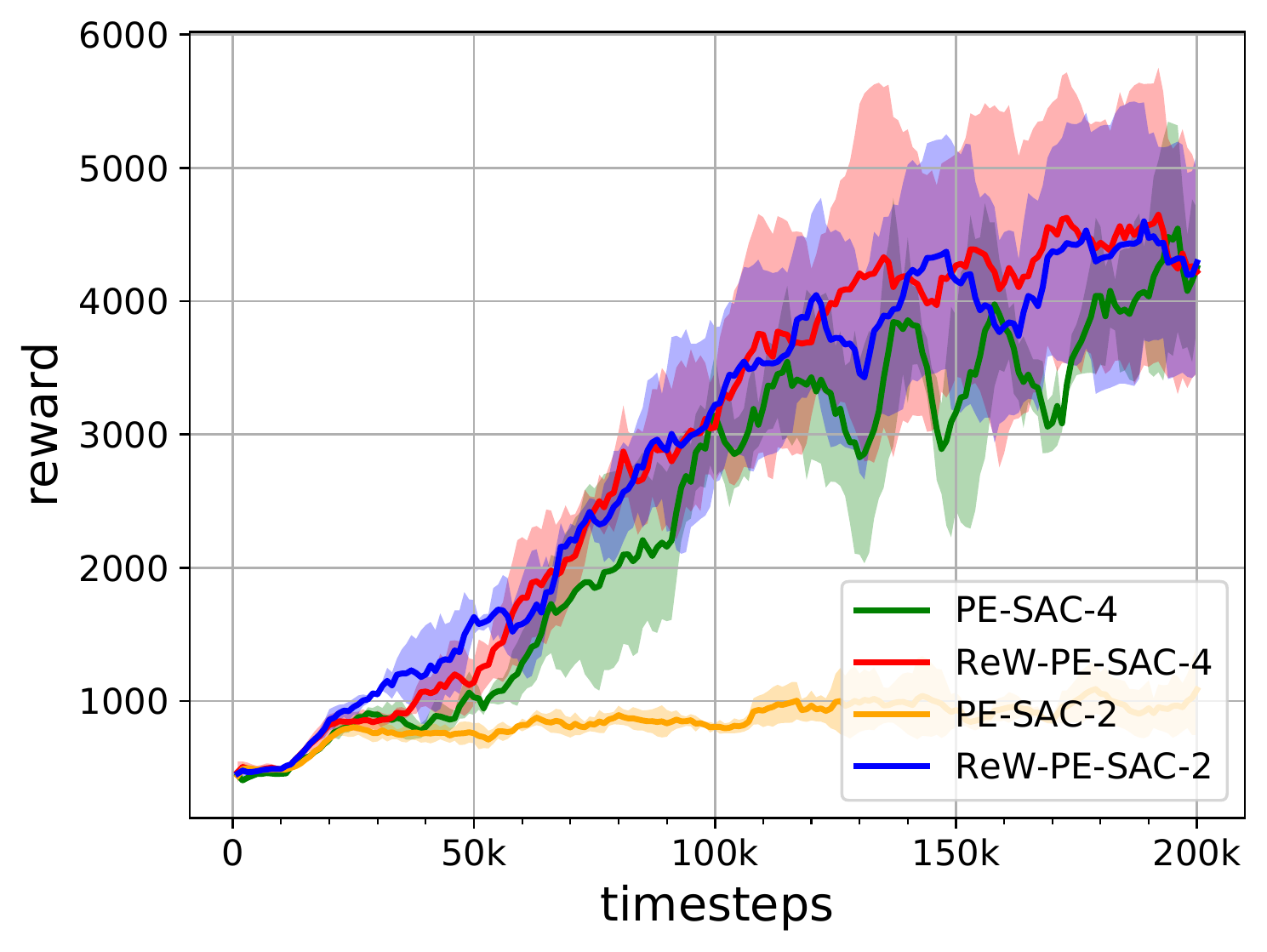}
	\caption{\label{fig:performance_of_different_model} Learning curves for PE-SAC and ReW-PE-SAC with the dynamics models using different numbers of hidden layers.}
\end{figure}

\begin{figure*}
	\begin{center}
		\subfigure[Overall Trend]{
			\centering
			\includegraphics[width=0.3\linewidth]{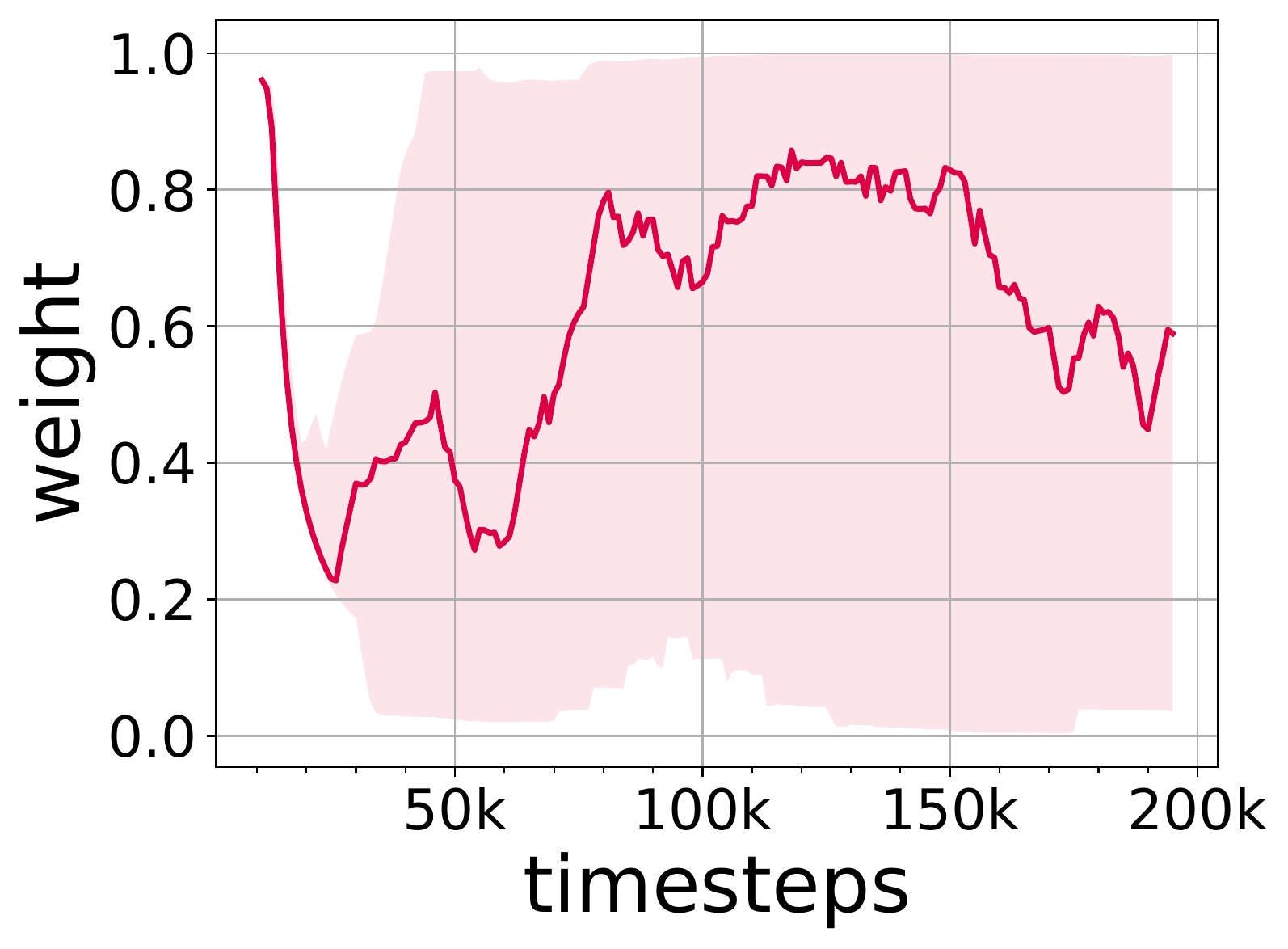}
			\label{fig:total_weights}
		}%
		\subfigure[Different Prediction Depth]{
			\centering
			\includegraphics[width=0.3\linewidth]{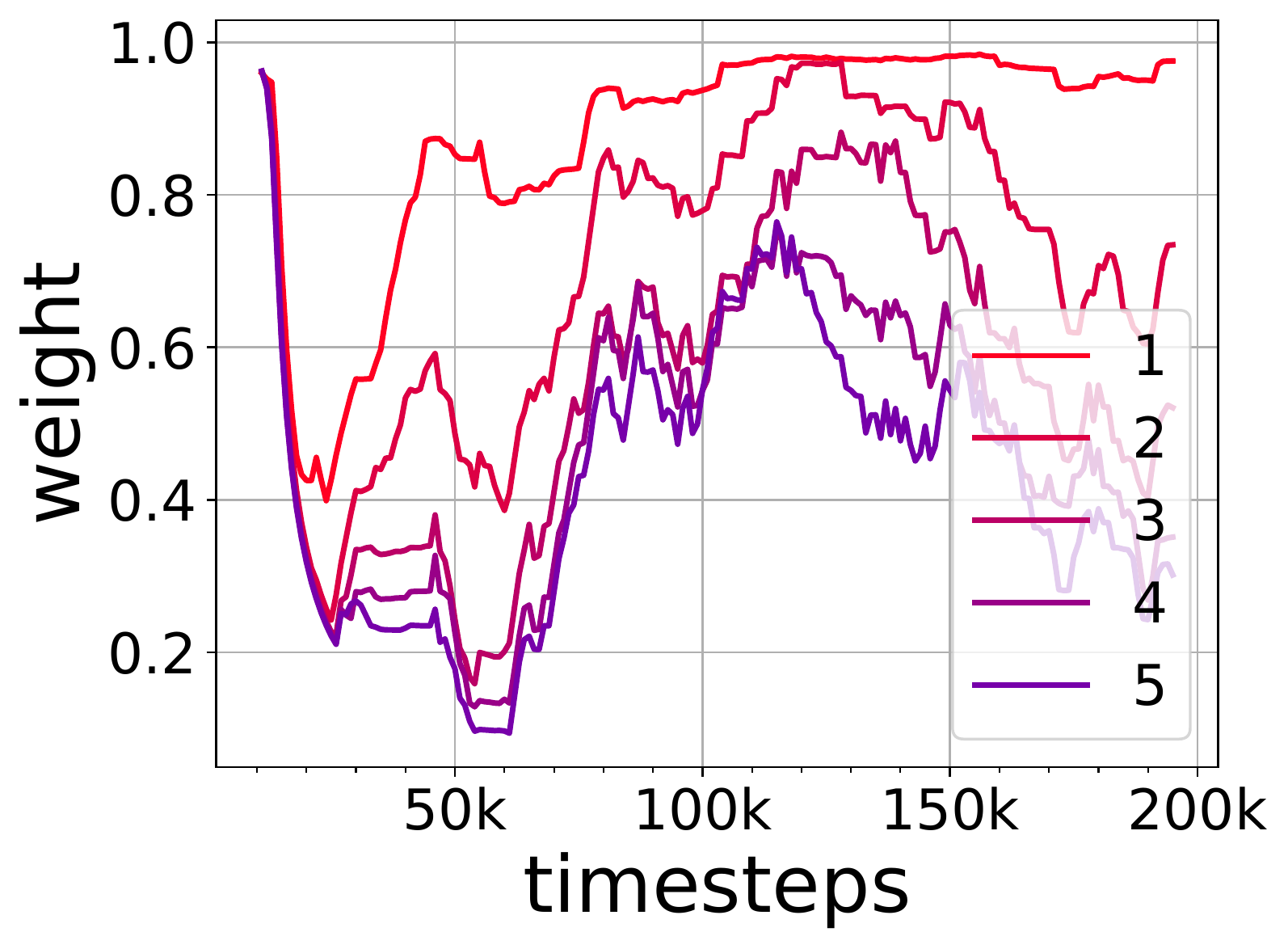}
			\label{fig:weights_in_fifferent_depth}
		}%
		\subfigure[Different Soft Scale]{
			\centering
			\includegraphics[width=0.3\linewidth]{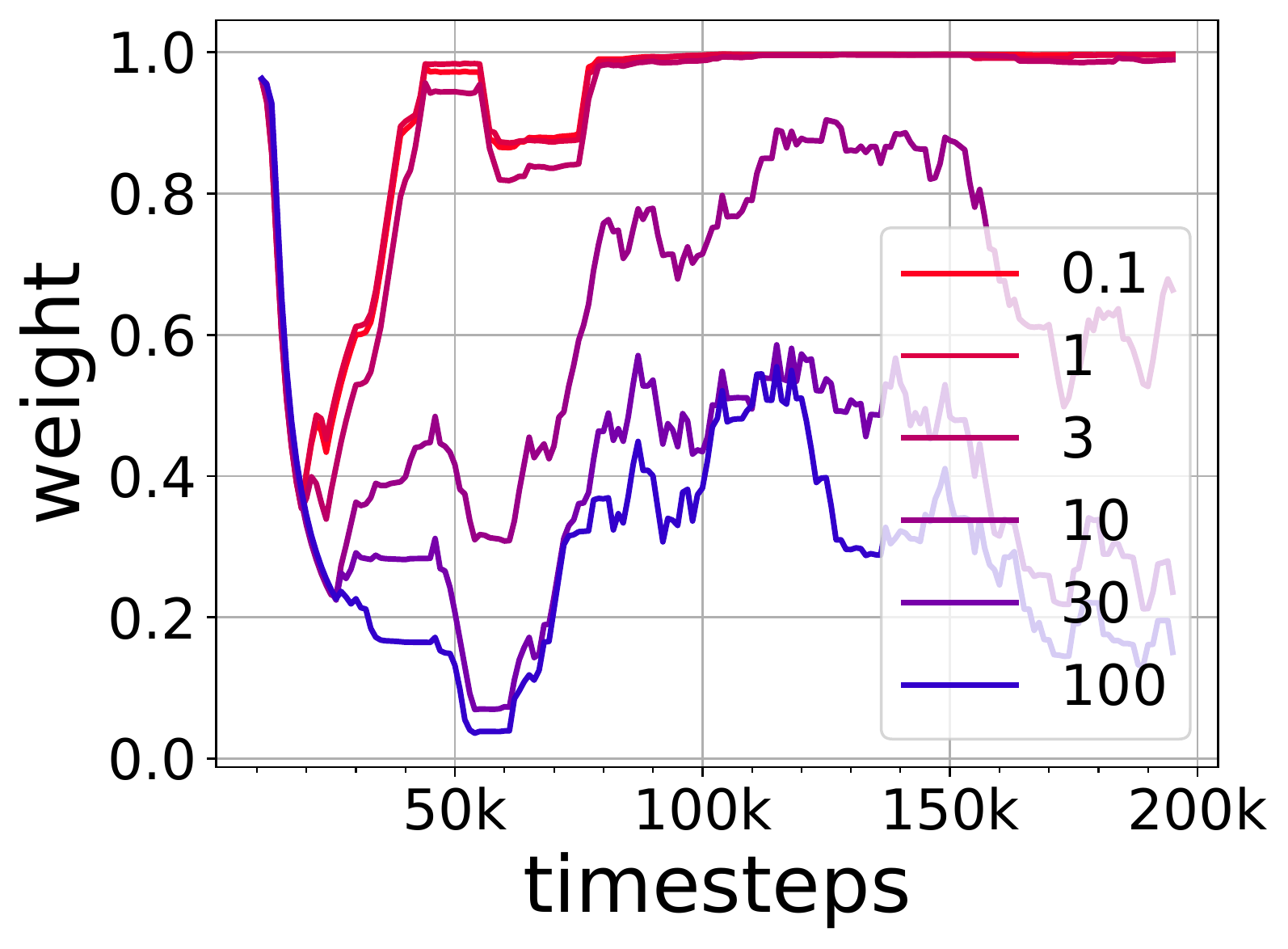}
			\label{fig:weights_in_fifferent_scale}
		}%
	\end{center}
	\caption{Predicted weights for different generated samples in different stages of training process.}
\end{figure*}

We construct the dynamics models with different prediction accuracy through adjusting the number of the hidden layers in them from $4$ to $2$.
We run the algorithms of PE-SAC and ReW-PE-SAC with these dynamics models on the tasks of Ant.
The learning curves of them are plotted in Figure~\ref{fig:performance_of_different_model}.

When the number of the hidden layers is decreased, the performances of PE-SAC drops significantly. 
This means that the dynamics models with $2$ hidden layers have strong negative effect on the training process.
The performances of ReW-PE-SAC remain roughly unchanged, which means that our method can effectively reduce the negative effect of the generated samples with prediction errors.
The above analysis gives a possible explanation for the phenomenon that ReW-PE-SAC has higher performance improvement on the more complex tasks, like Slimhumanoid and Walker2d.

\subsection{The Trend of the Predicted Weight}
\label{sec:weights}

In this section, we analyze the overall trend of the predicted weights and the relation between the weights and the prediction depth and the soft scale $\lambda_e$.
We run the algorithms of ReW-PE-SAC on the task of Swimmer with only $1$ random seed,
and record the predicted weights of generated samples at the first step of each episode. 
The predicted weights are changed with the process of training, so computing the average on different seeds is meaningless.
The 25 precent point, median and 75 precent point are plotted in Figure~\ref{fig:total_weights}.
Then, we split these weights according to the prediction depth, and plot the median of the weights of different prediction depth in Figure~\ref{fig:weights_in_fifferent_depth}.
Finally, we generate some extra data using different $\lambda_e\in\{0.1, 1, 3, 10, 30, 100\}$, and plot the median of predicted weights on them in Figure~\ref{fig:weights_in_fifferent_scale}.

In Figure~\ref{fig:total_weights}, the weights are lower in the earlier and later stages but are higher in the middle stage 
(The weight function's initial output is about $0.95$ as that the bias of last layer is initialized to $3.0$.).
The trend reflects the change of the accuracy of the dynamics model and the q-value and policy functions.
In the earlier stage, the dynamics model is imprecise, so most of the generated transitions are rejected.
Then, the weights become to increase as the improvement in the prediction precision of the dynamics model.
However, in the later stage, the precision of q-value function also improves, while the model has reached its bottleneck.
This results in the decline of the weights.
From Figure~\ref{fig:weights_in_fifferent_depth}, we find that the predicted weights decrease with the planning steps which accords with the fact that the prediction errors
accumulates with steps.
From Figure~\ref{fig:weights_in_fifferent_scale}, we also found that the weights decrease with the scale which is caused by the difference of the distributions of the actions in the training and predicting process of the dynamics model.
These phenomenons further verify that the learned weight function is reasonable.

\section{Conclusion}

In this paper, we have proposed a novel and efficient model-based reinforcement learning approach, which adaptively adjusts the weights of all generated transitions through training a weight function to reduce the potential negative effect of them.
We measure the effect of reweighted imaginary transitions through computing the difference of the losses computed on the real transitions before and after training with them, and minimize the difference to optimize the weight function by the chain rule.

Experimental results show that our method obtains the state-of-the-art performance on multiple complex continuous control tasks.
The learned weight function can provide reasonable weights for different generated samples in different stages of training process.
We believe that the weight function can be utilized to adjust some hyper-parameters, like planning horizon, in the future.

%

\section{ Acknowledgments}
This work is funded by the National Natural Science Foundation of China (Grand No. 61876181 No. 61673375 and No.61721004), Beijing Nova Program of Science and Technology under Grand No. Z191100001119043, the Youth Innovation Promotion Association, and CAS and the Projects of Chinese Academy of Science (Grant No. QYZDB-SSW-JSC006).

\bibliography{meta_reweight}

\end{document}